# SEMANTIC SENTIMENT ANALYSIS BASED ON PROBABILISTIC GRAPHICAL MODELS AND RECURRENT NEURAL NETWORKS

A Thesis Presented to the Department of Computer Science

African University of Science and Technology

In Partial Fulfilment of the Requirements for the Degree of

Master of Science

By

Osisiogu, Ukachi Oluwaseun

Abuja, Nigeria

July 2019

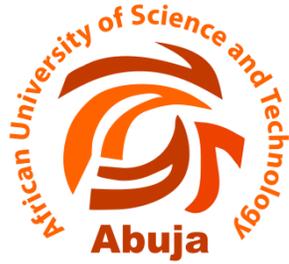

**African University of Science and Technology [AUST]**
*Knowledge is Freedom*

# APPROVAL BY

**Supervisor**

Surname: Odumuyiwa

First name: Victor

Signature:

**The Head of Department**

Surname: Rajesh

First name: Prasad

Signature:

**CERTIFICATION**

This is to certify that the thesis titled "Semantic Sentiment Analysis based on Probabilistic Graphical Models and Recurrent Neural Networks" submitted to the School of Postgraduate Studies, African University of Science and Technology (AUST), Abuja, Nigeria for the award of the Master's degree is a record of original research carried out by Osisiogu, Ukachi in the Department of Computer Science.

# ABSTRACT

Sentiment Analysis is the task of classifying documents based on the sentiments expressed in textual form, this can be achieved by using lexical and semantic methods. The purpose of this study is to investigate the use of semantics to perform sentiment analysis based on probabilistic graphical models and recurrent neural networks. In the empirical evaluation, the classification performance of the graphical models was compared with some traditional machine learning classifiers and a recurrent neural network. The datasets used for the experiments were IMDB movie reviews, Amazon Consumer Product reviews, and Twitter Review datasets. After this empirical study, we conclude that the inclusion of semantics for sentiment analysis tasks can greatly improve the performance of a classifier, as the semantic feature extraction methods reduce uncertainties in classification resulting in more accurate predictions.

## ACKNOWLEDGMENT

I would like to express my sincere gratitude to God for giving me the wisdom, knowledge, and understanding needed to complete this project.

I would also want to express my sincere thanks to my Supervisor Dr. Victor Odumuyiwa whose guidance and encouragement during this project work was of immense help to me.

I also want to thank Dr. Rajesh Prasad whose encouragement and feedback enabled me to stay on course.

**DEDICATION**

I dedicate this project to my parents and my sisters who never give up on me and always have believed me.

vi

**TABLE OF CONTENTS**

**LIST OF FIGURES**

**LIST OF TABLES**

# CHAPTER ONE

# INTRODUCTION

Sentiment analysis is the subject of natural language processing technique whose main aim is to perform the task of classifying, extracting and detecting attitudes, sentiments, and opinions of the different aspects or topics of an entity or product expressed in textual form. The usefulness of sentiment analysis includes but not limited to determining the level of consumer satisfaction (Ren & Quan, 2012), analyzing, political movements (Tumasjan, Sprenger, Sandner, & Welpe, 2010), performing market intelligence (Li & Li, 2013), measuring and improving brand reputation(Zhang et al., 2013), box office prediction (Nagamma, Pruthvi, Nisha, & Shwetha, 2015), and many others (Nasraoui, 2008), (Ravi & Ravi, 2015).

## 1.1 Research Background

Access to people's opinions, sentiments and evaluations have increased in general and in a wide variety of fields in e-commerce (Akter & Wamba, 2016), tourism (Alaei, Becken, & Stantic, 2017), and social networks (Sehgal & Agarwal, 2018). One of the major causes of this is the rise of Big Data. Consumers now read product reviews by previous customers, with this the improvement of products and services carried out by service providers is enhanced through feedback obtained by customers through channels that employ textual data.

## 1.2 Problem Statement

Despite the stated usefulness and advantages that come with sentiment analysis, there are a lot of challenges. These challenges include: the usage of sarcastic statements especially in social network platforms like Twitter; the possibility of words possessing different meanings  - for instance, a word can bear positive meanings in some contexts, and negative in another; people also express their opinions in varied ways so a small change in the syntax of the message communicated can mean something different in the implied opinion. Also, some of the opinions expressed cannot be categorized as a particular type of sentiment, since sometimes they may appear to be subjective and also appear neutral in another perspective. Also, issues like this could raise questions like "at what point can we classify a statement as being neutral or positive or negative?" The aforementioned shows us how challenging sentiment analysis can be even for humans.

Problems in sentiment analysis can be addressed by using a variety of methods. Some of these techniques are known as Probabilistic Graphical Models such as Bayesian Networks, Hidden Markov Models and Conditional Random Fields. In this review section, we focus on Hidden Markov Models (HMM) which is also known as a variation of Dynamic Bayesian Network. HMM is a modeling technique that can observe the state of a sequence and also assign transition probabilities to perform classifications on the sequence (Jurafsky & Martin, 2019).

## 1.3  Aim and Objectives

This project aims to investigate the use of probabilistic graphical models and RNN to perform semantic sentiment analysis of textual data. Sentiment Analysis can be formulated as a text classification problem.

Specific objectives include:

1. Verifying if the semantic representation of data can further inform the classification process of an algorithm. The focus on probabilistic graphical models (PGMs) are emphasized because of their ability to model dependencies between events.
2. Carrying out performance evaluation of graphical models, traditional machine learning algorithms and RNN in sentiment analysis tasks on some benchmark datasets.

## 1.4  Project Outline

The outline of this project is well designed to establish the concept of semantics when performing text classification tasks. In Chapter 2, an extensive review of the previous works done in the use of graphical models for text classifications tasks is carried out; then some other machine learning algorithms used to demonstrate the application of semantics and non-semantics are briefly discussed. In Chapter 3 the methodologies executed is explained as a systematic approach to the investigation made in this project is discussed. Then in Chapter 4, empirical experiments are carried out showing the results and giving reasons why the results are obtained. In Chapter 5, recommendations for future work are given and the conclusion of findings reiterated.

# CHAPTER TWO

# LITERATURE REVIEW

This section presents an extensive review of the use of Probabilistic Graphical Models (PGMs) for sentiment analysis tasks and other text classification problems. A focus on two graphical models will be carried out - Hidden Markov Models and Bayesian Network Classifiers. Some traditional machine learning algorithms like Logistic Regression, Naïve Bayes Classifier, Random Forest Classifier, Support Vector Machine, and Decision Tree will be discussed. The chapter will be concluded with a brief discussion of Recurrent Neural Networks (RNN).

## 2.1    Graphical Models

### 2.1.1   Bayesian network classifiers

Sentiment analysis problems can be approached through a PGM known as a Bayesian Network; Bayesian networks are modeling techniques that allow for the description of dependency relationships between different variables by the application of a directed graph structure that encodes conditional probability distributions (Grosan & Abraham, 2011). By storing expert knowledge in the structure of these models Bayesian Probabilistic models can perform or support classification tasks (Guti, Bekios-calfa, & Keith, 2019). It is worthy to note that even though Naïve Bayes corresponds to the simplest Bayesian Network model – a simple Bayesian Network with a single root node[35]. Also, a large number of works in the field apply this technique to perform the classification task, and thus this technique has been well studied and extensively used in sentiment analysis literature. For this reason, Naïve Bayes has not been considered in this review. However, a focus on more complex Bayesian Network structures and approaches that have been applied to sentiment analysis are given more consideration. Based upon the exposition made by Crina and Ajith (Grosan & Abraham, 2011) and amongst other references, a brief introduction into the concept of Bayesian Networks (BNs) is provided.

Following the context of modeling and machine learning problems, Bayesian networks are normally used to find relationships among a large number of words. Thus, BNs provide an adequate tool used to model these relationships. BNs consists of a directed acyclic graph where each node represents a random variable and the edges between the nodes represent an influence relationship. Conditional probability distributions are typically used to model these influences.

*"A Bayesian Network for a set of variables consists of a network structure that encodes conditional independence assertions about the variables and a set of local probability distributions associated with each variable. Together these two components allow a defining joint probability distribution for the set of all problem variables. This conditional independence allows building a compact representation."*(Grosan & Abraham, 2011)

To define conditional probability distributions a table known as Conditional Probability Table (CPT) is given. This table assigns probabilities to the variable node depending on the values of its parents in the graph. In the case where a node does not have a parent, the CPT assigns a probability distribution to that random variable[33]. Figure 1 shows an example BN and its corresponding CPT

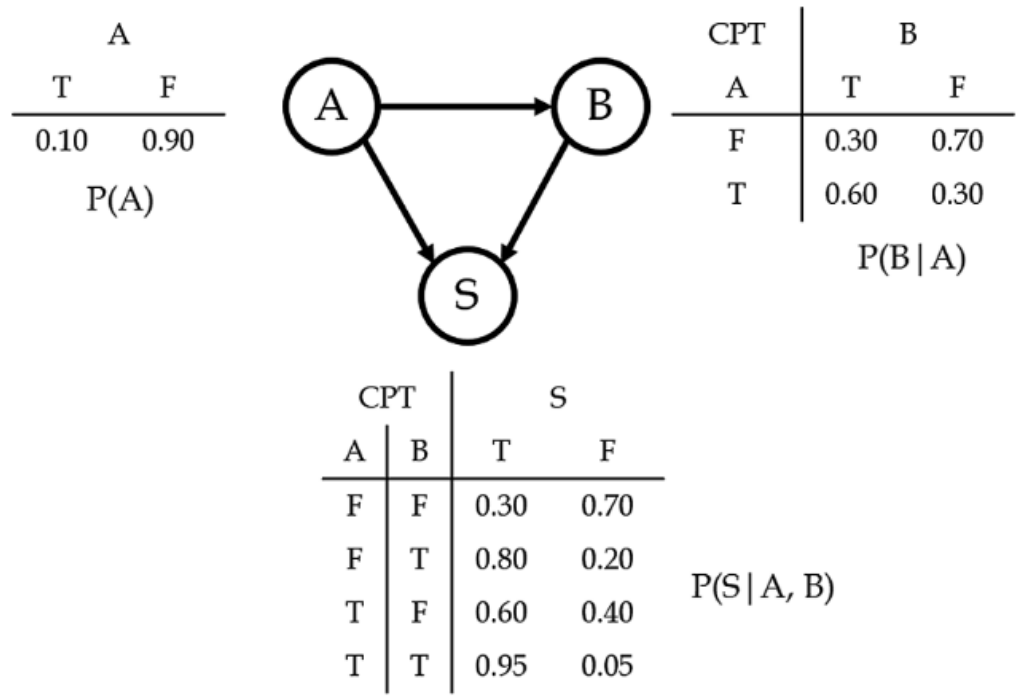

Figure 1: Example of a simple Bayesian Network

To build a classifier using Bayesian Network, it is required that the structure of the network is first learned along with their respective CPTs. Furthermore, the fundamental concept of CPTs can be extended to the continuous case in which the variables can base on the other laws of probability such as Gaussian Distribution or solved by applying discretization (John & Langley, 2013),(Driver & Morrell, 1995),(Friedman & Goldszmidt, 1996)

Although inference in any Bayesian Network is an NP-hard problem(Gregory F Cooper, 1990), there are efficient alternatives that exploit conditional independence for some types of networks (Heckerman, 2008). Also, one of the benefits of Bayesian Networks in their ability to directly handle incomplete datasets if one of their entries are missing.

Wan's work (Wan & Gao, 2016) and an article by Al-Smadi et al. (Al-Smadi, Al-Ayyoub, Jararweh, & Qawasmeh, 2019) show that one of the recurrent use of Bayesian Networks is classification as they are directly used as a sentiment classifier in these works and they obtained competitive results and in some cases higher when compared with other approaches.

In another work proposed by Chen et al. (Chen et al., 2011) a parallel algorithm for the structure learning of large-scale text datasets for Bayesian networks was created. With the application of a MapReduce cluster, dependencies between words were captured. This approach allows for obtaining a vocabulary for extracting sentiments. Experiments were carried out using a blog's dataset; this work points out that features can be extracted despite fewer predictor variables.

Lane et al. (Lane, Clarke, & Hender, 2012) addressed issues facing most sentiment analysis tasks such as choosing the right model, feature extraction and dealing with unbalanced data. Although the main task is classification, they took into consideration two different techniques. Firstly, the classification subjectivity and then the polarity determination. In this work, several techniques to extracting features were evaluated, as dealing with unbalanced data was considered before training. It turns out that the Bayesian Network model tested showed a decrease in their performance when applying data balancing techniques. This behavior was different from that of the other classifiers

Ortigosa et al. (Ortigosa-Hernández et al., 2012) approach a multidimensional problem for the sentiment analysis task. Here they were able to use a three-dimensional related feature for sentiment analysis. This was most useful in cases where a one-dimensional approach was not suitable. Furthermore, they proposed a network of multi-dimensional Bayesian classifiers (De Waal & Van Der Gaag, 2007), (Bielza, Li, & Larrañaga, 2011) and applied semi-supervised techniques to avoid the manual labeling of examples.

A two-stage Markov Blanket Classifier was proposed by Airoldi et al. (Airoldi, Bai, & Padman, 2006) and Bai (Bai, Padman, & Airoldi, 2004) to perform extraction of sentiments from unstructured text, such as film reviews, using BNs. In their approach, a Tabu Search algorithm (Pardalos, Du, & Graham, 2013) is used to prune the resulting network to obtain more accurate classification results. Although this helps to prevent overfitting their work does not efficiently exploit dependencies among sentiments.

In contrast, Olubolu (Sylvester Olubolu Orimaye, 2013) proposed improvement for the Bayesian Network classification model that fully exploits sentiment dependencies by including sentiment-dependent penalties for scoring functions of Bayesian Networks (e.g. K2, Entropy, MDL, and BDeu). This proposed modification derives the dependency structure of sentiments using conditional mutual information between each pair of variables in the dataset. In (S.O. Orimaye, Pang, & Setiawan, 2016) the knowledge contained in SentiWordNet was evaluated. The experimental results obtained showed that this sentiment-dependent model could improve the classification accuracy in some domains.

A hierarchical approach for the modeling of simple and complex emotions in the text is proposed by Ren and Kang (Ren & Kang, 2013). Many documents are associated with complex human emotions that are a mixture of simple emotions and they are difficult to model using traditional machine learning techniques (Naïve Bayes, and Support Vector Machine) that were used as baselines in this work. The traditional machine learning algorithms were able to model texts with simple emotions while the hierarchical methods were more suitable for modeling documents with complex emotions. The analysis performed in this work also points out that there is a relationship between the topics of documents and the emotions contained in them.

In another study by Wang et al. (L. Wang, Ren, & Miao, 2016) also attempts to address the challenges in complex emotions. Here the author evaluates multilabel sentiment analysis techniques on a dataset obtained from Chinese weblogs – Ren CECps. Utilizing the theory of probabilistic graphical models and Bayesian Networks, the latent variables that represent emotions and topics are used to realize complex emotions from the sentences of weblogs. Further analysis carried out in this project also demonstrates the effectiveness of the model in distinguishing the polarity of emotions a domain.

Chaturvedi et al. (Chaturvedi, Cambria, Poria, & Bajpai, n.d.) proposed a BN that is used in conjunction with a Convolutional Neural Network for the detection of subjectivity. In this work, the authors introduce a Bayesian Deep Convolutional Neural Network that possesses the ability to model higher-order features through several sentences in a document. They utilized Gaussian Bayesian Networks to learn the features that are fed to the convolutional neural network. Their proposal delivers superior results when compared to the algorithms used as baselines in the project.

### 2.1.2 Hidden Markov Models

Before delving into the discussions on the applications of HMMs to sentiment analysis problems, a brief introduction to the Hidden Markov Models using as a foundation for the exposition made by Jurafsky and Martin (Jurafsky & Martin, 2019) is carried out.

The HMM is based on amplifying the Markov Chain. A Markov Chain – the model that provides information about the probabilities of sequences of random variable states, each of which can take on values from some set. These sets can be words, or tags and even symbols that represent a variety of things. A Markov chain is useful when we need to compute a probability for a sequence of observable events. In several cases the events of interest are hidden - they are not observed directly. HMMs makes it possible to observe hidden and observable events. They are modeled as causal factors in our probabilistic model. HMM is a generative probabilistic model which consists of N not directly observable hidden states:

$$S = \{S_1, S_2, S_3 \ldots S_N\} \quad (1)$$

The emission alphabet i.e. M distinct observation symbols per state is given by:

$$V = \{V_1, V_2, V_3 \ldots V_M\} \quad (2)$$

Let the current state at time t, be denoted as $q^{(t)}$. Let the state transition probability distribution for the HMM be denoted as $A = \{a_{ij}\}$ and is given by

$$a_{ij} = P[q^{(t+1)} = S_j | q^{(t)} = S_i > 0] \; for \; 1 \leq i \leq N, 1 \leq j \leq N \quad (3)$$

Furthermore, an observation symbol probability distribution B $\{b_{jk}\}$ in state $S_j$ is given by

$$b_{jk} = P[v_k | q^{(t)} = S_j \wedge v_k = O_t \; for \; 1 \leq j \leq N, 1 \leq k \leq M \quad (4)$$

Then an initial probability distribution over states $\pi = \{\pi_i\}$ with

$$\pi_i = P[q^{(1)} = S_i for 1 \leq i \leq N] \quad (5)$$

With all the above parameters, the Hidden Markov model can be written as a 3-tuple

$$\lambda = (A, B, \pi) \quad (6)$$

As a consequence, there is a bit of resemblance between a doubly stochastic process and an HMM where after each period t, the system can remain in its current state or make a transition. Given the state $S = \{S_1, S_2, S_3 \dots S_N\}$, the system can get to another state in a single step implemented by $a_{ij} > 0$ in (3). To give a visual example, in Figure 2 a stochastic process with two states $S = \{S_1, S_2\}$ is shown. To integrate another stochastic process each of these states will now emit a symbol $O_t$ of the emission symbol $V$ at time $t$; then the model generates a sequence of observable states.

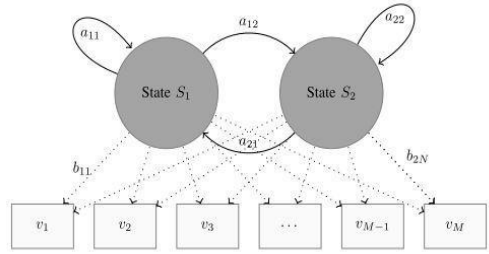

*Figure 2: The emission probabilities for a HMMs with two states and emission of M symbols at time t*

$$O(T) = [O_1, O_2, \ldots O_T] \qquad (7)$$

Until a time, step T.

As stated by Rabiner (RABINER & L., 1993) hidden Markov models should be marked by three fundamental tasks:

***Task 1 (Evaluation):*** Estimating the probability of an observation sequence $O(T) = [O_1, O_2, \ldots O_T]$ for a model $\lambda = (A, B, \pi)$ this can be executed with the Forward algorithm.

***Task 2: (Decoding):*** Estimation of the corresponding state sequence $Q = [q^{(1)}, q^{(2)} \ldots q^{(t)}]$ to an observation $O(T)$ and a model λ. This can be done with the Viterbi algorithm.

***Task 3: (Learning):*** Fine-tuning of the parameters

$\lambda = (A, B, \pi)$ to maximize $P(O(T) \vee \lambda)$ via a method named Baum-Welch algorithm.

*Table 1: Articles based on the use of HMMs for sentiment analysis categorized by year*

| Year | Articles |
|------|----------|
| 2013 | (Klinger & Cimiano, 2013),(Rustamov, Mustafayev, & Clements, 2013; Zhang et al., 2013) |
| 2014 | (West, Paskov, Leskovec, & Potts, 2014; Yin, Han, Huang, & Kumar, 2014; Zhang et al., 2014) |
| 2015 | (Liu et al., 2015; Pröllochs, Feuerriegel, & Neumann, 2015; Xie, Jiang, Ye, & Li, 2015) |
| 2016 | (Wei & Yongxin, 2016) |
| 2017 | (Suleiman, Awajan, & Al Etaiwi, 2017),(Kim, Kim, & Lee, 2017) |
| 2018 | (Zhao & Ohsawa, 2018), (Kang, Ahn, & Lee, 2018) |

## 2.1.2.1 A systematic review of the applications of HMMs for sentiment analysis

In this section, the details of the review carried out are discussed. Relevant examples of how the reviewed works have carried out sentiment analysis tasks will be discussed; we also demonstrate how these articles were selected.

## 2.1.2.2 Methodology

To gather related publications for this review, primary computer science, machine learning, and natural language processing publication databases were considered: IEEE Xplore, Scopus, Advanced Computational Linguistics, (ACL) Digital Library and Google Scholar. A keyword search was performed to gather publications relevant to this review; examples of keywords used are "Hidden Markov Model", "Sentiment Analysis", "Probabilistic Graphical Models", and "Semantic Sentiment Analysis". The returned articles were narrowed down to fit the subject of discussion and relevant use cases. Each paper reviewed was based on the following criteria. (1) Does the paper present an implementation of HMM to carry out sentiment analysis or text classification? (2) Does the work present evaluations that compare the performance of the proposed model with other algorithms. (3) Does the paper contain suggestions for future research? After examining the search results closely, a total of 14 publications were reviewed as shown in Table 1.

### 2.1.2.3   Reviewed Works

Kang et al. (Kang et al., 2018) focused on the sequences of words to address some of the issues faced with the use of lexicons when performing sentiment analysis. They propose the use of a model that will focus on word orders without the need for extracting sentiment lexicons. To achieve this an ensemble of text-based HMM is proposed. This model employed the boosting and clustering of words produced by latent semantic analysis.

After the input data has been labeled and words in the textual data have been clustered, the ensemble is used to create a classifier. The sentences were categorized into positive (Y=1) and negative (Y=−1) sentences. Based on the sentence orientation the corresponding labels were assigned using this equation

$$\hat{Y} = argmax \{f_{Y=1}(s), f_{Y=-1}\} \qquad (8)$$

The above process is known as TextHMMs when repeated over severally in different ways, an ensemble of TextHMMs is created – with this approach multiple and diverse patterns were reflected in the training data. To summarize, the authors were able to implement this approach with the following steps (1) Employ a clustering algorithm based on Latent Semantic Analysis to determine the hidden states of words (2) Construct the Hidden Markov Model (3) Through boosting build an ensemble of TextHMMs.

The method of the ensemble method used for the TextHMMs is known as Boosting. When tested on different datasets ranging from movie reviews to opinions of competing for products on the web and compared with other state-of-the-art machine learning algorithms used in SA such as Conditional Random Fields (CRF) -a dependency tree-based method, a matrix-vector recursive neural network (MV-RNN) model, a convolutional neural network using pre-trained vectors from Word2Vec and a multinomial naïve Bayes-Support Vector machines with uni-bigrams (NBSVM), the work shows that the accuracy of the Ensemble-HMM is generally higher than the compared methods in most of the cases. Further, this model was tested with real-life datasets; the results produced were relatively impressive.

Another implementation of HMMs was carried out by Zhao and Ohsawa in (Zhao & Ohsawa, 2018) but this time in a higher dimension. This work utilized a 2dHMM – HMMs which provides us with the ability to model sentiment analysis in higher dimensions. The SA experiments carried out in this research were not based on the subjectivity of the user alone but also considered the behavior of the user. This paper shows that the proposed model can capture the event where the sentiment of a user concerning a product is influenced by the observation of the last two reviews or top-rated review by the user. With this, one could be able to model dependencies between the last two comments and a top-rated review and the words of a user to perform sentiment classification. This paper was able to demonstrate this by using a 2dHMM to model such a situation given the web page of a product. Here the two latest reviews and a top-rated review of a Japanese Tea product were considered in the classification of the sentiment. It turns out that the 2dHMM produced better precision and F1 scores when compared to other conventional machine learning algorithms that perform classification based on the user's textual data alone. The reason for better performance is based on the ability of the 2dHMM to model dependencies between recent reviews, top-rated reviews, and user's sentiment.

Another approach to the use of HMM to perform SA is proposed in (Kim et al., 2017). This technique learns patterns of word sequences and sentimental word transitions. The HMM deployed was trained on constructed informative hidden states and transition patterns of words in sentimental sentences. Syntactic-sentimental (positive-adjectives) features were created by combining syntactic (adjectives, adverbs, etc.) and sentimental (positive, negative, neutral) features. GMMs (Gaussian Mixture Models) were applied to the produced unigrams to get the SIGs (Similar syntactic and sentimental Information Groups). SIGs transitions between words were modeled. SIGs are employed as hidden states of the HMMs and can model transition patterns seen in sentimental sentences. The performance evaluation carried out with Health Care Reform (HCR) dataset with about 839 tweets shows that HMMs with 4 states in most cases outperformed HMM algorithms with lower states. Other conventional machine learning methods like the SVM, NB and algorithms were implemented in these works(da Silva, Hruschka, & Hruschka, 2014), (Speriosu, Sudan, Upadhyay, & Baldridge, 2011) & (Saif, He, & Alani, 2012).

In performing sentiment analysis for other languages other than English in (Suleiman et al., 2017), the review is given by Suleiman et. al, it was observed that text classification tasks were possible. With a degree of modification and by further research the model can be adapted to sentiment analysis tasks. Also, Wei and Yongxin (Wei & Yongxin, 2016) demonstrate the use of HMM to perform network public sentiment analysis of Chinese text by using conventional methods. They pointed out that because of the robustness that HMM provides, it can be used to describe the probability model of the stochastic process's statistical properties. After carrying out performance evaluation using Chinese textual data, the work showed that the HMM outperformed the Naïve Bayes and Support Vector Machine algorithms.

Liu et al. (Liu et al., 2015) used a self-adaptive HMM to perform emotion classification on Weibo a microblogging website in China (similar to Tweeter). Due to the text mining challenges like word segmentation, the arrival of new words on Weibo and ambiguity, a self-adaptive HMM was proposed. The application of HMMs facilitated the development of a more fine-grained emotional analysis, the creation of useful features to be trained on the HMM and a further improvement of HMM (self-adaptive HMM) to mine textual data from Sina-Weibo. The implementation of self-adaptive HMM required features extracted from textual data to be modeled as observed variables; in this approach, states are considered to be a set of values that represent an emotional category. The self-adaptive mechanism is obtained through the parameter estimation made by the Particle Swarm Optimization algorithm (PSO). The experimental result demonstrates that HMM outperforms SVM and NB when classifying certain emotions.

Nicolas et al. (Pröllochs et al., 2015) demonstrated that improvements can be made on the sentiment analysis of financial news with the detection of negation scopes. To achieve this, both rule-based algorithms and the Hidden Markov Models were employed. First, data preprocessing tasks involving cleaning, tokenization, the removal of stop words, parts-of-speech tagging, and stemming were carried out. Afterward, rule-based detection of negation scopes was carried out via the application of linguistic rules. In this work, the HMM-Based Detection of Negation scope was employed to align with domain-specific features and peculiarities of a chosen area. To utilize HMM for the prediction of negation scope directly observable states are given, the actual words are then chosen as emission symbols, then word stems acts as comparisons. The performance of the variants of HMMs (supervised and unsupervised) was then carried out. In the evaluation of the polarity of news announcements an approach known as the Net-optimism sentiment measure (Henry, 2008), (Demers & Vega, 2008), was applied. Using a manually labeled dataset that consists of 400 extracted sentences the HMM implementations were evaluated. The model when used to predict negation scope performed below the baseline. In other words, the rule-based approach performed better with significant results in the negation scope forecast.

In another work (Xie et al., 2015), a variation of HMM known as selective HMM was used to perform financial trend predictions with Twitter Moods; this was carried to achieve high prediction performance and gain good control over the financial trend prediction. First, the Twitter moods are evaluated and extracted by building a sentiment lexicon based on a profile of mood states (POMS) Bipolar and WordNet to effectively extract six-dimensional society moods from enormous tweets; to determine which of the Twitter Moods possess the most predictive power, the Granger causality analysis (GCA)(Gruber, Rosen-Zvi, & Weiss, 2007) between the financial index and the Twitter mood is carried out to determine the most important mood that facilitates the prediction of the market trend. It was then discovered that the Twitter mood had the most predictive power. WordNet Synsets were used to expand the Bipolar POMS lexicon; Yifu et. al points out that selective HMMs are based on a concept known as a selective prediction - a prediction framework that can characterize the results of its predictions. With selective HMM, financial index and Twitter moods are combined. Also, during the training of the selective HMM, the MapReduce framework was employed for efficient evaluation. For the evaluation of the proposed algorithm, two Twitter datasets and two financial data were used in the experiments. The proposed model produced the least error margin when compared with other algorithms.

Kunpeng et al. (Zhang et al., 2013) proposed a probabilistic graphical model that can represent relationships between social brands and users. The model can collectively measure reputations of entities in social networks; it not only captures network information but also includes the semantic information from users in terms of the comments they make. To achieve this the model adopts a block-based Markov Chain Monte Carlo (MCMC) sampling method to deduce the probability of hidden variables, user positives, and brand reputations. This technique was used to avoid the computational complexity that comes with the direct calculation of the joint probability of the hidden variables due to large state space. One of the vital advantages of this model is its ability to reduce the biased effect from a single user and a single comment as this most times occur in other conventional methods used in performing sentiment analysis. Experiments were conducted using a large amount of data from Facebook taking into consideration relevant and unbiased features. This was compared with existing ranking systems - the IMDB movie ranking and top business school ranking by the US News and World report; the correlation between these ranking systems and the proposed algorithms were significant.

In another work done by Kunpeng et al. (Zhang et al., 2014) a rather robust method of implementation was employed with conditional random fields (CRF). Factors that influence sentiment were considered, for instance, the newly emerged internet language, emoticons, positive words, negative words, negation words and also useful information about the sentence structure like conjunction words and comparisons. Additionally, they also utilized context information to capture the relationship among sentences for the improvement of document-level sentiment classification; and incorporated human interaction to improve sentiment identification accuracy and construct a large training set.

The input of the algorithm includes specified subjects and a set of corresponding documents, while the output of the algorithm will assign a sentiment value to a particular sentence in a document. To capture context information among sentences in a document to predict their sentiment class; a form of sequence labeling is employed. The model is used to assign a label to each sentence as it corresponds to a certain sentence sequence. In this work the CRF employed provides a probabilistic framework for calculating the probability of a random variable or vector over corresponding label sequences (Y) conditioned on a random variable/vector over sequence data to be labeled (X). In this work, a linear chain structure was employed (a structure similar to a Hidden Markov model). The document containing multiple sentences served as the observation sequence, this was used as the given condition for the label sequence-tagged as a label.

After which log-likelihood technique was used to perform parameter estimation, and then many iterative scaling algorithms were used to optimize the parameters. The Viterbi algorithm was employed to make an inference with an approach very similar to the forward-backward algorithm of an HMM. Semantic features (number of positive/negative words, positive or negative emoticons, comparative sentences, type of conjunction words) and Syntactic features (sentence positions, simple or compound sentence, the position of positive/negative words, the position of negation words, comparison subject, similarity to neighboring sentences) were combined to form semantic-syntactic features and used with the CRF algorithm. The evaluation of the model shows that the CRF based model when compared with other methods - the compositional semantic rule (CSR)(a rule-based algorithm), Support Vector Machine (SVM), Logistic Regression (LR) and Hidden Markov Model (HMM) and tested on datasets collated from Amazon reviews, outperforms the other methods by 5-15%. However, the CSR performs best on the Facebook comments dataset and the rest of the methods produce similar results.

In (Rustamov et al., 2013) the parameters of the HMM were estimated according to the corresponding classes and trained using the Baum-Welch algorithm; these estimations along with a scaled-forward algorithm were used to provide the probabilities associated with a given review and its corresponding class. After that, these calculated probabilities were passed through a decision-making block to perform the final classification task. It turns out that the HMM with three states gives the best results when compared with predictions made by the Fuzzy Control System (FCS), and Neuro-Fuzzy Models (ANFIS).

Also, this work proposed a hybrid system that combines and HMMs and the other two algorithms to obtain better results.

Peculiar with most probabilistic graphical models, it can be observed that modeling dependencies between parameters can improve accuracy in sentiment classification. The authors of (Yin et al., 2014) demonstrate this further. With the Dependency-Sentiment-LDA, sentiment classification was carried out, however, the text was modeled in the form of a Markov Chain (Gruber et al., 2007) to facilitate this purpose. After evaluation, it was observed that the introduction of topic dependencies and sentiment prior information increased the accuracy in sentiment classification.

West et al. (West et al., 2014) developed a graphical model that can synthesize network and linguistic information to make better predictions about these parameters. This also demonstrates the ability for graphical models to perform multidimensional sentiment analysis. The idea employed in this work was to predict A's sentiment of B using the synthesis of the structural context around A and B inside a social network and sentiment analysis of the evaluative texts relating A to B. Although an NP-hard problem, the authors point out an approach that can relax the problem to an efficiently solvable hinge-loss Markov Random field (a variation of HMM). In this paper, they further demonstrated how joint models of text and network structure can excel where their components part cannot.

In the bid to analyze and quantify to what extent joint probabilistic models outperform pipeline probabilistic models in terms of extraction of aspects, subjective phrases and the relation between them, Klinger and Cimiano (Klinger & Cimiano, 2013) discovered that the use of a joint inference model made way for the yielding of a deeper and fine-grained analysis of sentiments by modeling the relation between aspects and subjective phrases; and also outperforms the pipeline model in the prediction of aspects. However, in the prediction of subjective phrases and relations, the pipeline model outperforms the joint model. The inference on the imperatively-defined factor graphs (the employed probabilistic graphical model) is carried out by the Markov Chain Monte Carlo Technique. The data representation employed possess peculiar similarities with the data model of a Hidden Markov Model; templates were used to define the sets of variables that form the graphical structure of the probabilistic model, the features that lead to the factor's score and the parameters associated with them. Then effective sampling strategies were employed for the joint and pipeline models coupled with adequate objective functions and training.

## 2.2 Traditional Machine Learning Algorithms

Some machine learning algorithms have performed reasonably well in text classification tasks. In this project, we define them as baseline models for the empirical evaluation of PGMs. A brief discussion of these algorithms is made in this section

### 2.2.1 Logistic Regression

One of the foremost methods of text classification algorithms is known as Logistic Regression (LR). This algorithm was introduced and developed by a statistician known as David Cox (Snell, 2018) LR is a linear classifier with a decision boundary defined by $\theta^T x = 0$ and it predicts probabilities rather than classes (Fan, Chang, Hsieh, Wang, & Lin, 2008)(Genkin, Lewis, & Madigan, 2007). To define a class, it takes the maximum value of the predicted probability of the respective class. However, there are certain limitations to this algorithm; LR classifiers work well for predicting categorical outcomes. To ensure optimal performance, the prediction requires that each data point be independent identically distributed (iid) to perform best. These data points attempt to predict the outcomes based on a set of independent variables(Guerin, 2016)

### 2.2.2 Naïve Bayes Classifier (NBC)

Naïve Bayes Classification has been widely used for text classification tasks that involve document categorization tasks (Kaufmann, 1969). The Naïve Bayes method is based on Bayes theorem, formulated by Thomas Bayes (Pearson, 1925). Information retrieval systems have widely adopted this algorithm (Qu et al., 2018). This technique is a generative model – a traditional method of text categorization. In this project, we apply the Naïve Bayes classifier on textual data that has its feature extracted by the TF-IDF approach. One peculiar limitation of the NB classifiers is its inability to work on unbalanced classes. Also, this classifier makes a strong assumption about the shape and dependencies of data distribution (Y. Wang, Khardon, & Protopapas, 2012). NBC is also limited by data scarcity; for any value in the feature space, a likelihood value must be estimated by a frequentist (Ranjan, 2017).

## 2.2.3 Support Vector Machine (SVM)

Vapnik and Chervonenkis (Vapnik & Chervonenkis, 1964) developed the original version of SVM in 1963. However, B.E. Boser et al. (Bregni et al., 2005) made modifications to this version to suit into a non-linear form in the early 1990s. Although the SVM was designed for binary classifications, many researchers work on multi-class problems using this technique. Some of the limitations of Support Vector Machines especially in text classification tasks stems from the lack of transparency in results caused by a large number of dimensions.

## 2.2.4 Decision Trees (DT)

Another important classification algorithm for text and data mining is the decision tree [69]. Decision Tree classifiers have been successfully used in varied areas of classification. It was introduced as a classification tool by D.Morgan(Magerman, 1995) and inductions developed by J.R. Quinlan(Quinlan, 1986). This technique employs a hierarchical composition of the data space. The main idea behind this algorithm is found upon the creation of a tree based on the attribute for categorized data points. A major challenge in the implementation of a decision tree is in the assignment of attributes to the parents' level or the child level. To tackle this problem statistical modeling for feature selection in trees was created. Although the decision tree is a fast algorithm for both learning and prediction. It is also extremely sensitive to small perturbations in the data (Giovanelli, Liu, Sierla, Vyatkin, & Ichise, 2017), and can easily overfit (Quinlan, 1987). Mitigation effects on these problems can be obtained through pruning and validation

methods(Giovanelli et al., 2017). This algorithm also possess out of sample predictions problems (Jasim, 2016)

### 2.2.5 Random Forest (RF)

One of the ensemble learning methods that is mainly used in text classification tasks is known as Random Forests or Random Decision Forests technique. This technique was introduced by T.Kam Ho in 1995 (Kowsari et al., 2019). The decision trees generate random decision trees that are trained and predictions are assigned by voting. Some of the limitations of Random Forest remain that they are quite slow to create predictions once trained. However, they possess a better speed of convergence when compared with other machine learning algorithms. To achieve faster prediction results the number of trees in the forest must be reduced, this can result in lesser time complexity in the prediction step.

### 2.3 Recurrent Neural Network (RNN)

Neural Networks are designed to learn a multi-connection of layers that every single layer only receives the connection from the previous and provides connections only to the next layer in a hidden part. In the context of text classification tasks, the input layer may be constructed via TF-IDF, word embeddings or some feature extraction approach. The output layer may consist of the number of classes for multi-classification or only one neuron for binary classification.

An important variation of this that has been utilized by several researchers for text mining and classification tasks is the recurrent neural network (RNN) (Sutskever, Martens, & Hinton, 2011). The RNN assigns more weights to the previous data points of a sequence. Thus, this feature makes the RNN a powerful approach to sequential data, text, and strings. RNNs consider the information of previous nodes in a very sophisticated method which allows for better semantic analysis of a data set's structure. RNN implementations are carried out through LSTMs or GRUs for text classification. The input layer contains word embeddings, the other parts of this neural network possess hidden layers and finally output layer.

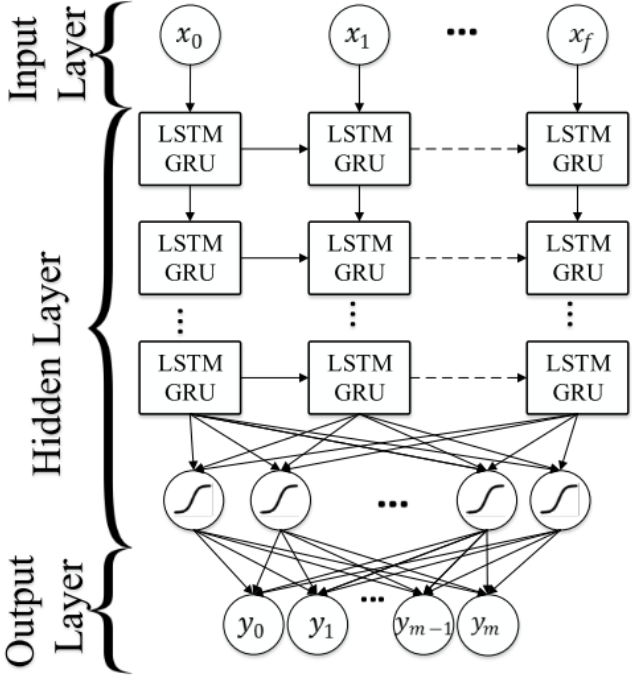

*Figure 3: Diagram showing the architecture of an LSTM and GRU neural network.* Source: ((Sutskever et al., 2011).

### 2.3.1 Long Short-Term Memory (LSTM)

S. Hochreiter and J. Schmidhuber (Hochreiter & Schmidhuber, 1997), first introduced the LSTM, ever since this architecture has been augmented by many research scientists. LSTM is a special type of RNN that addresses the problem of vanishing gradients by preserving long term dependencies more effectively when compared to the basic RNN. LSTM possess a chain-like structure similar to RNN, LSTM utilizes multiple gates to carefully regulate the degree of information that is allowed into each node state. A form of bias can be introduced into RNNs when later words are more influential than earlier ones. This, however, can be resolved with the deployment of max-pooling areas.

This review points out some of the major text classification machine learning algorithms that have been used to perform sentiment analysis. A focus on sentiment analysis tasks based on Probabilistic Graphical models was made and a brief discussion on the other machine learning algorithms that have produced amazing results was made. This review shows the peculiar nature of the PGMs as they exercise certain features that are worthy of note. This project aims to investigate the impact of semantics in sentiment analysis tasks and text classification in general. Based on this extensive review, we demonstrate that the ability of graphical models to model dependencies between words implies a level of semantic application. Performing empirical experiments to prove this claim will further be a contribution to the body of knowledge in this field.

**CHAPTER THREE**

**METHODOLOGY**

The scope of this project lies within the investigation of semantic representation and semantic feature extractions of textual data for the use of sentiment analysis - a text classification problem. We aim to discover how Probabilistic Graphical Models (PGMs) can encode the semantic representation of textual data by the establishment of dependencies between words. To carry a proper investigation an empirical methodology is implemented. This methodology was chosen amongst others because it provides the platform to compare the sentiment classification performance of existing classification algorithms that encode semantics with classification algorithms that are non-semantic. These comparisons will be validated with the help of baseline performance indicators. The results of this project for the sentiment classification tasks aim to point out that the existing dependencies formulated by the graphical models are semantic and therefore can perform better than non-semantic approaches for the same task.

## 3.1 Datasets

The datasets used in the empirical research were IMDB movie reviews, Amazon Product reviews, and Twitter datasets. These datasets were chosen as it spans across the most common datasets used for the analysis of sentiments.

### 3.1.1. IMDB Dataset

This is a dataset for binary sentiment classification containing 25,000 highly polar movie reviews for training, and 25,000 for testing (Maas et al., 2011).

### 3.1.2. Amazon Product Review

This dataset is a subset of the main dataset that contains product reviews and metadata from Amazon, including 142.8 million reviews spanning May 1996 - July 2014. This dataset includes reviews (ratings, text, helpfulness votes), product metadata (descriptions, category information, price, brand, and image features), and links (also viewed/also bought graphs). For this project, we use only about 28,000 datasets and performed some resampling methods where necessary.

### 3.1.3. Twitter Datasets

This dataset consists of 4,242 tweets manually labeled with their polarity.

The experiment carried out in this project were carried to investigate:

1. The use of graphical models to carry out sentiment analysis tasks and,

2. The use of semantic sentiment analysis and non-semantic methods as shown in Table 2.

*Table 2: Outline of experiments carried out.*

| Method | Algorithm | Textual Representation |
|---|---|---|
| Graphical Models | Bayesian Network | TF-IDF (Word Vectors - Sparse Vector Representation) |
| Non-Semantic | Logistic Regression, Support Vector Machine, Naïve Bayes, Decision Trees, Random Forest | TF-IDF (Sparse Vector Representation) |
| Semantic Representation | Long Short-Term Memory | Word Embeddings (GloVe, Word2Vec) (Dense Vector Representation) |

**CHAPTER FOUR**

**EXPERIMENTATION, RESULTS AND DISCUSSIONS**

The workflow employed in this project was designed in such a way to derive a level of accuracy that measures up to the conventional standards in text classification problems. The Bayesian Network used in this experiment were obtained from Weka (Hall et al., 2009); the implementation of the traditional machine learning classifiers (Logistic Regression, Support Vector Machine, Naïve Bayes, Decision Trees, Random Forest) used in this project was obtained from the SciKit Learn API (Buitinck et al., 2013) built for machine learning. TensorFlow (Martın et al., 2005) was used to implement the implementation of the neural network of the text classification algorithms. Figure 4 gives a diagrammatic representation of the process.

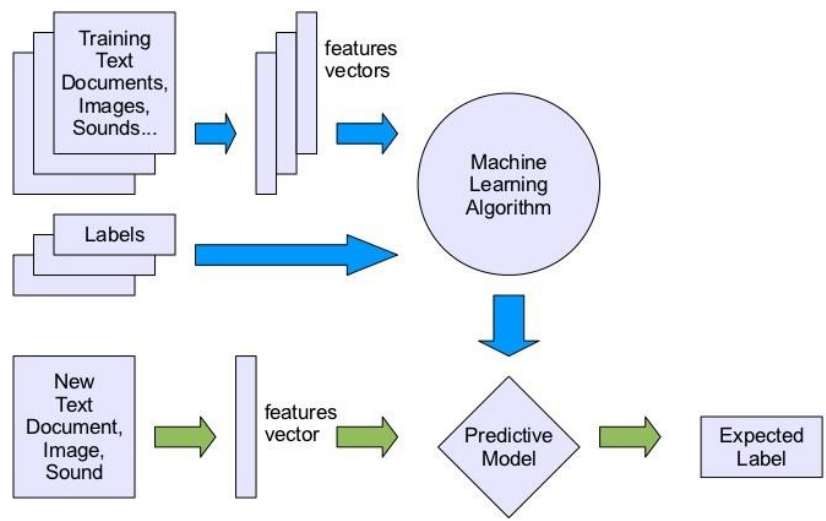

*Figure 4: Workflow of experimentation executed for the text classification task.*

Firstly, for all cases (for all the algorithms) text preprocessing will take place. This text preprocessing as will be discussed in the later section will be different for each classifier while in some cases the text will be represented as feature vectors (TF-IDF), in some other cases they will be represented as Word Embeddings. This form of various representations is fed into their respective algorithms to train the models. After the model has been trained. New text documents (test set) are fed into the trained model and of course, are also represented in the same way as the test dataset. With this, the predictive models make predictions that are checked against the actual results of the test dataset to measure the performance of the model.

## 4.1   Graphical Models

In this experiment, we center on the use of different scoring functions and search algorithms in the Bayesian Network (BN). This was done to investigate just how well these algorithms can perform Sentiment Classification (SC) tasks and to determine how the inclusion of semantics can help inform the classification of the algorithms.

Given a data $D = \{y_1, \ldots y_N\}$, to find the BN that best fits the data, a scoring function is used. Let this scoring function be represented as $\emptyset$. The problem of learning a Bayesian Network can thus be defined like this:

Given data, $D = \{y_1, \ldots y_N\}$ and a scoring function $\emptyset$, we find a BN $B \in B_n$ that maximizes the value $\emptyset(B, T)$.

However, in finding an approximate solution, Cooper (Gregory F Cooper, 1990) showed that computing the inference of a general BN is an NP-hard problem. To circumvent this problem, some researchers worked on finding an approximate solution to this. However, Dagum and Luby (Dagum & Luby, 1993) also showed that finding an approximate solution is NP-hard. Further work showed that if the search space can be restricted the computational requirement of finding the solution of the BN can be reduced. In contrast, Chickering (Chickering, 1996) showed that learning the structure of a BN is NP-hard even for networks constrained to have in-degree of at most 2.

Dasgupta (Dasgupta, 1999) even showed that learning 2-polytress is also NP-hard. Due to the hardness of these results exact polynomial-time bounded approaches for learning BNs have been restricted to tree structures. Therefore, the conventional methodologies for addressing the problem of learning the BNs became a heuristic search, based on scoring metrics optimization, conducted over some search space. Some of the search space includes:

- Network structures
- Equivalence classes of network structures
- Orderings over the network variables

Search algorithms that can be used to search the space are:

- Greedy hill-climbing
- Simulated annealing

- Genetic Search algorithm
- Tabu search

The scoring metrics optimizations can be further classified into two:

### 4.1.1 Bayesian Scoring Functions

This computes the posterior probability distribution, starting from a prior probability distribution on the possible networks, conditioned to data, $D\ i.e.\ P(B|D)$

### 4.1.2 Information-theoretic scoring functions

The score of a Bayesian network B is related to the compression that can be achieved over the data D, with an optimal code induced by B.

In this project, we implemented variations of search algorithms and corresponding score metrics using Weka. With Weka, the search algorithms (cf. section 4.1.4) and scoring functions (cf. Table 4) were implemented and tested with the datasets employed in this project.

Some Useful Notations to aid in the definitions of the scoring functions are presented in Table 3, while the definition of the scoring functions used in the experimentation is shown in Table 4.

*Table 3: Definitions of useful Notations required to aid the definitions of the scoring functions*

| Notation | Definition |
|---|---|
| $B_s$ | Network Structure |
| $r_i$ | Number of states of the finite random variable $X_i$ |
| $x_{ik}$ | $k$-th value of $X_i$ |
| $q_i = \prod_{X_j \in \prod_{X_i}} r_j$ | Number of possible configurations of the parent set $\prod_{X_i}$ of $X_i$ |
| $w_{ij}$ | $j$-th configuration of $\prod_{X_i}$ $(1 \leq j \leq q_i)$ |
| $N_{ijk}$ | Number of instances in the data $D$ where the variable $X_i$ takes its $k$-th value $x_{ik}$ and the variables in $\prod_{X_i}$ take their $j$-th configuration $w_{ij}$ |
| $N_{ij} = \sum_{k=1}^{r_i} N_{ijk}$ | Number of instances in the data $D$ where the variables in $\prod_{X_i}$ take their $j$-th configuration $w_{ij}$ |

| $N_{ik} = \sum_{j=1}^{q_i} N_{ijk}$ | Number of instances in the data $D$ where the variable $X_i$ takes its $k$-th value $x_{ik}$ |
|---|---|
| $N$ | Total number of instances in the data $D$ |

### 4.1.3 Scoring Functions used in the experiment

*Table 4: Definitions of scoring functions used in the experiment*

| Scoring Function | Description | Category |
|---|---|---|
| Bayesian Metric | The Bayesian metric of a Bayesian network structure $B_s$ for a database D is $$Q_{Bayes}(B_s, D) = P(B_s) \prod_{i=0}^{n} \prod_{j=1}^{q_i} \frac{\Gamma(N'_{ij})}{N'_{ij} + N_{ij}} \prod_{k=1}^{r_i} \frac{\Gamma N'_{ij} + N_{ijk}}{\Gamma(N'_{ijk})}$$ | Bayesian |
| BDeu | The Bayesian Dirichlet (BD) metric of a Bayesian network structure $B_s$ for a database D is | Bayesian |
|   | $$Q_{BDeu}(B_s, D) = \log(P(B_s))$$ $$+ \sum_{i=1}^{n} \sum_{j=1}^{q_i} \left( \log\left( \frac{\Gamma(N'_{ij})}{N'_{ij} + N_{ij}} \right) \right.$$ $$\left. + \sum_{k=1}^{r_i} \log\left( \frac{\Gamma N'_{ij} + N_{ijk}}{\Gamma(N'_{ijk})} \right) \right)$$ This appears when $$P(X_i = x_{ik}, \pi_{X_i} = w_{ij}|G) = \frac{1}{r_i q_i}$$ |   |
| K2 | The K2 metric of a Bayesian network structure $B_s$ for a database D is $$Q_{K2}(B_s, D) = \log(P(B_s))$$ $$+ \sum_{i=1}^{n} \sum_{j=1}^{q_i} \left( \log\left( \frac{(r_i - 1)!}{(N_{ij} + r_i - 1)!} \right) \right.$$ $$\left. + \sum_{k=1}^{r_i} \log(N_{ijk}!) \right)$$ | Bayesian |

| MDL | MDL (Minimum Description Length) metric $Q_{MDL}(B_s, D)$ of a Bayesian network structure $B_s$ for a database (the data) $D$ is $$Q_{MDL}(B_s, D) = H(B_s, D) + \frac{K}{2} \log N$$ | Information-theoretic |
|---|---|---|
| Entropy | Entropy metric, $H(B_s, D)$ of a network structure and database is defined as $$H(B_s, D) = -N \sum_{i=1}^{n} \sum_{j=1}^{q_i} \sum_{k=1}^{r_i} \frac{N_{ijk}}{N} \log \frac{N_{ijk}}{N_{ij}}$$ And the number of parameters $K$ as $$K = \sum_{i=1}^{n} (r_i - 1) \cdot q_i$$ | Information-theoretic |
| AIC | AIC metric $Q_{AIC}(B_s, D)$ of a Bayesian network structure $B_s$ for a database (the data) $D$ is $$Q_{AIC}(B_s, D) = H(B_s, D) + K$$ | Information-theoretic |

### 4.1.4 Search algorithms used in this experiment

- **K2**: This Bayesian Network learning algorithm uses the hill-climbing technique restricted by an order of variables. It can perform this by adding arcs with a fixed ordering of the variables present in a dataset (G.F. Cooper & Herskovits, 1990). In the implementation of this algorithm, the K2 operation can either perform random ordering of the nodes made at the beginning of the search or perform the ordering of the nodes using the dataset.

- **Hill Climbing** (Buntine, 1996): Another Bayesian Network learning algorithm that utilizes the hill-climbing algorithm to add, reverse or delete arcs with no fixed ordering of variables.

- **Repeated Hill Climber**: This search algorithm starts with a randomly generated network. It repeatedly applies the hill climber to reach a local optimum and then returns the best structure of the various runs.

- **LAGD Hill Climber**: This is another variation of hill climbing that performs hill climbing with a look ahead on a limited set of best scoring steps.

- **Tabu Search**: This Bayesian Network algorithm uses a tabu search (Bouckaert, 1995) to find a well scoring Bayesian Network structure.

- **Tree Augmented Naïve Bayes (TAN)** : (Cheng & Greiner, 2013), (Nir, Geiger, & Goldszmidt, 2014) is formed by calculating the maximum weight spanning tree using the Chow and Liu algorithm (C.K.Chow & Liu, 1968). It returns a Naive Bayes network augmented with a tree.

### 4.1.5 Data Preparation

The datasets were prepared according to the WEKA's ARFF format by concatenating the negative and positive reviews for each dataset and created a string data file in the ARFF format. The string data file was then preprocessed using the *weka.filters.unsupervised.attribute.StringToWordVector* package. This package converted the string data file to a TFIDF data file in the ARFF format. The TFIDF format as earlier discussed is a numerical representation of the text variables that are supported by the *Bayes* package. It is worthy of note that this representation still maintains the dependency relationship between words (variables) as in the original string format.

Table 5 shows the number of attributes used. This was carefully selected after testing a range of attributes, the number of attributes that resulted in having the best performance was then selected.

*Table 5: Distribution of prepared datasets used in WEKA.*

| Dataset | Instances | Negative/Positive | Attributes |
| --- | --- | --- | --- |
| IMDB | 50000 | 25000/25000 | 5000 |
| Amazon | 28332 | 8435/19897 | 2500 |
| Twitter | 4438 | 2218/2218 | 65 |

### 4.1.6 Results

The Bayesian Network algorithm was implemented using the *weka.classifiers.Bayes* package of the WEKA data mining framework.

Using different search algorithms and carefully selected scoring functions, the following results were obtained.

*Table 6: Precision, Recall and F1 score of the respective Bayes classifier when applied to the IMDB movie review dataset.*

| IMDB Dataset | | | | |
|---|---|---|---|---|
| Search Algorithm | Scoring Function | Precision | Recall | F1 Score |
| K2 | Bayes | 0.857 | 0.857 | 0.857 |
| | BDeu | 0.857 | 0.857 | 0.857 |
| | MDL | 0.857 | 0.857 | 0.857 |
| | Entropy | 0.857 | 0.857 | 0.857 |
| | AIC | 0.857 | 0.857 | 0.857 |
| EBMC | K2 | 0.776 | 0.774 | 0.774 |
| Hill Climber | Bayes | 0.857 | 0.857 | 0.857 |
| | BDeu | 0.857 | 0.857 | 0.857 |
| | MDL | 0.857 | 0.857 | 0.857 |
| | Entropy | 0.857 | 0.857 | 0.857 |
| | AIC | 0.857 | 0.857 | 0.857 |
| | Bayes | 0.857 | 0.857 | 0.857 |
| LAGD Hill Climber | BDeu | 0.857 | 0.857 | 0.857 |
| | MDL | 0.857 | 0.857 | 0.857 |
| | Entropy | 0.857 | 0.857 | 0.857 |
| | AIC | 0.857 | 0.857 | 0.857 |
| Repeated Hill Climber | Bayes | 0.857 | 0.857 | 0.857 |
| | BDeu | 0.857 | 0.857 | 0.857 |
| | MDL | 0.857 | 0.857 | 0.857 |
| | Entropy | 0.857 | 0.857 | 0.857 |
| | AIC | 0.857 | 0.857 | 0.857 |
| Tabu Search | Bayes | 0.857 | 0.857 | 0.857 |
| | BDeu | 0.857 | 0.857 | 0.857 |
| | MDL | 0.857 | 0.857 | 0.857 |
| | Entropy | 0.857 | 0.857 | 0.857 |
| | AIC | 0.857 | 0.857 | 0.857 |
| TAN | Bayes | 0.858 | 0.858 | **0.858** |
| | BDeu | 0.858 | 0.858 | **0.858** |
| | MDL | 0.858 | 0.858 | **0.858** |
| | Entropy | 0.858 | 0.858 | **0.858** |

|  | AIC | 0.858 | 0.858 | **0.858** |

Table 7: Precision, Recall and F1 score of the respective Bayes classifier when applied to the Amazon product review dataset.

| Amazon Dataset | | | | |
|---|---|---|---|---|
| Search Algorithm | Scoring Function | Precision | Recall | F1 Score |
| K2 | Bayes | 0.730 | 0.747 | 0.738 |
|  | BDeu | 0.730 | 0.747 | 0.738 |
|  | MDL | 0.730 | 0.747 | 0.738 |
|  | Entropy | 0.730 | 0.747 | 0.738 |
|  | AIC | 0.730 | 0.747 | 0.738 |
| Hill Climber | Bayes | 0.730 | 0.747 | 0.738 |
|  | BDeu | 0.730 | 0.747 | 0.738 |
|  | MDL | 0.730 | 0.747 | 0.738 |
|  | Entropy | 0.730 | 0.747 | 0.738 |
|  | AIC | 0.730 | 0.747 | 0.738 |
| LAGD Hill Climber | Bayes | 0.730 | 0.747 | 0.738 |
|  | BDeu | 0.730 | 0.747 | 0.738 |
|  | MDL | 0.730 | 0.747 | 0.738 |
|  | Entropy | 0.730 | 0.747 | 0.738 |
|  | AIC | 0.730 | 0.747 | 0.738 |
| Repeated Hill Climber | Bayes | 0.730 | 0.747 | 0.738 |
| | BDeu | 0.730 | 0.747 | 0.738 |
| | MDL | 0.730 | 0.747 | 0.738 |
| | Entropy | 0.730 | 0.747 | 0.738 |
| | AIC | 0.730 | 0.747 | 0.738 |
| Tabu Search | Bayes | 0.730 | 0.747 | 0.738 |
| | BDeu | 0.730 | 0.747 | 0.738 |
| | MDL | 0.730 | 0.747 | 0.738 |
| | Entropy | 0.730 | 0.747 | 0.738 |
| | AIC | 0.730 | 0.747 | 0.738 |
| TAN | Bayes | 0.735 | 0.750 | **0.742** |
| | BDeu | 0.735 | 0.750 | **0.742** |
| | MDL | 0.735 | 0.750 | **0.742** |
| | Entropy | 0.735 | 0.750 | **0.742** |
| | AIC | 0.735 | 0.750 | **0.742** |

*Table 8: Precision, Recall and F1 score of the respective Bayes classifier when applied to the Twitter dataset.*

| Twitter Dataset | | | | |
|---|---|---|---|---|
| Search Algorithm | Scoring Function | Precision | Recall | F1 Score |
| K2 | Bayes | 0.683 | 0.644 | 0.663 |

|  | BDeu | 0.683 | 0.644 | 0.663 |
|---|---|---|---|---|
|  | MDL | 0.683 | 0.644 | 0.663 |
|  | Entropy | 0.683 | 0.644 | 0.663 |
|  | AIC | 0.683 | 0.644 | 0.663 |
| EBMC | K2 | 0.685 | 0.646 | **0.665** |
| Hill Climber | Bayes | 0.683 | 0.644 | 0.663 |
|  | BDeu | 0.683 | 0.644 | 0.663 |
|  | MDL | 0.683 | 0.644 | 0.663 |
|  | Entropy | 0.683 | 0.644 | 0.663 |
|  | AIC | 0.683 | 0.644 | 0.663 |
| LAGD Hill Climber | Bayes | 0.683 | 0.644 | 0.663 |
|  | BDeu | 0.683 | 0.644 | 0.663 |
|  | MDL | 0.683 | 0.644 | 0.663 |
|  | Entropy | 0.683 | 0.644 | 0.663 |
|  | AIC | 0.683 | 0.644 | 0.663 |
| Repeated Hill Climber | Bayes | 0.683 | 0.644 | 0.663 |
|  | BDeu | 0.683 | 0.644 | 0.663 |

|  | MDL | 0.683 | 0.644 | 0.663 |
|---|---|---|---|---|
|  | Entropy | 0.683 | 0.644 | 0.663 |
|  | AIC | 0.683 | 0.644 | 0.663 |
| Tabu Search | Bayes | 0.683 | 0.644 | 0.663 |
|  | BDeu | 0.683 | 0.644 | 0.663 |
|  | MDL | 0.683 | 0.644 | 0.663 |
|  | Entropy | 0.683 | 0.644 | 0.663 |
|  | AIC | 0.683 | 0.644 | 0.663 |
| TAN | Bayes | 0.685 | 0.646 | **0.665** |
|  | BDeu | 0.685 | 0.646 | **0.665** |
|  | MDL | 0.683 | 0.644 | **0.663** |
|  | Entropy | 0.685 | 0.646 | **0.665** |
|  | AIC | 0.685 | 0.646 | **0.665** |

### 4.1.7 Discussion of Results

Table 6,7 and 8 shows the performance of the Bayesian Network classifier when applied to the IMDB movie review, Amazon Product review and Twitter dataset respectively. For each search algorithm, different scoring algorithms were used and the results obtained were the same except for the TAN search algorithm which consistently obtained a slightly

better result. The IMDB movie review has the highest amount of accuracy mainly because it has the largest amount of dataset when compared to the other datasets. For the Amazon product review dataset, the resulting precision and recall of most traditional machine learning algorithms would be affected because of the imbalanced data. The results obtained show that the imbalanced nature of the dataset causes a little effect in the resulting precision and recall figures.

## 4.2     Machine Learning Classifiers

### 4.2.1  Data Preparation

For each dataset each of the reviews was preprocessed using Python in the following steps:

- Removed punctuations
- Converted URLs to string "URL"
- Removed numbers and symbols to obtain alphanumeric data
- Coerced string to lowercase
- Used the *sklearn.feature_extraction* package to apply the TF-IDF vectorizer.

### 4.2.2  Results

To compare semantic and non-semantic methods various classifiers were implemented we use 20% of the training dataset for the validation of our model to check against overfitting.

*Table 9: Results obtained from the machine learning classifiers using the IMDB dataset.*

| IMDB Dataset | | | | |
|---|---|---|---|---|
| Classifier | Method of Evaluation | Precision | Recall | F1 Score |
| Logistic Regression (LR) | Micro average | 0.8969 | 0.8969 | 0.8969 |
| | Macro average | 0.8972 | 0.8968 | 0.8969 |
| | Weighted average | 0.8972 | 0.8969 | 0.8969 |
| Support Vector Machine (SVM) | Micro average | 0.8999 | 0.8999 | 0.8999 |
| | Macro average | 0.9001 | 0.8998 | 0.8999 |
| | Weighted average | 0.9001 | 0.8999 | 0.8999 |
| Naïve Bayes (NB) | Micro average | 0.8596 | 0.8596 | 0.8596 |
| | Macro average | 0.8607 | 0.8598 | 0.8596 |

|  | Weighted average | 0.8609 | 0.8596 | 0.8595 |
| --- | --- | --- | --- | --- |
| Decision Trees (DT) | Micro average | 0.7168 | 0.7168 | 0.7168 |
|  | Macro average | 0.7168 | 0.7168 | 0.7168 |
|  | Weighted average | 0.7168 | 0.7168 | 0.7168 |
| Random Forest (RF) | Micro average | 0.7414 | 0.7414 | 0.7414 |
|  | Macro average | 0.7463 | 0.7417 | 0.7403 |
|  | Weighted average | 0.7465 | 0.7414 | 0.7402 |

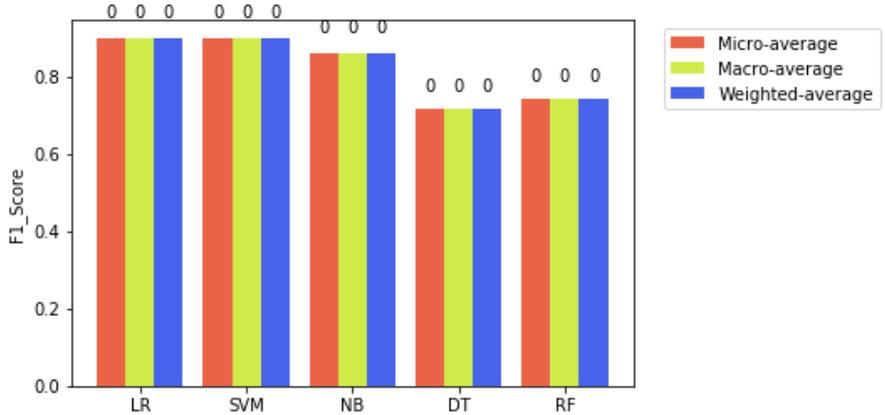

*Figure 5: Classification performance of machine learning classifiers on the IMDB datasets.*

*Table 10: Results obtained from the machine learning classifiers using the Amazon Product Review dataset.*

| Amazon Product Review Dataset (with Resampling) | | | | |
|---|---|---|---|---|
| Classifier | | Precision | Recall | F1 Score |
| Naïve Bayes (NB) | Micro average | 0.8464 | 0.8464 | 0.8464 |
| | Macro average | 0.6220 | 0.7976 | 0.6514 |
| | Weighted average | 0.9315 | 0.8464 | 0.8775 |
| Support Vector Machine | Micro average | 0.8872 | 0.8872 | 0.8872 |
| | Macro average | 0.6351 | 0.7276 | 0.6641 |
| | Weighted average | 0.9217 | 0.8872 | 0.9015 |
| Logistic Regression (LR) | Micro average | 0.8722 | 0.8722 | 0.8722 |
| | Macro average | 0.6364 | 0.7831 | 0.6713 |

|  | Weighted average | 0.9300 | 0.8722 | 0.8942 |
| --- | --- | --- | --- | --- |
| Decision Trees (DT) | Micro average | 0.8783 | 0.8783 | 0.8783 |
|  | Macro average | 0.5902 | 0.6329 | 0.6050 |
|  | Weighted average | 0.9039 | 0.8783 | 0.8899 |
| Random Forest (RF) | Micro average | 0.9298 | 0.9298 | 0.9298 |
|  | Macro average | 0.6792 | 0.5636 | 0.5872 |
|  | Weighted average | 0.9068 | 0.9298 | 0.9133 |

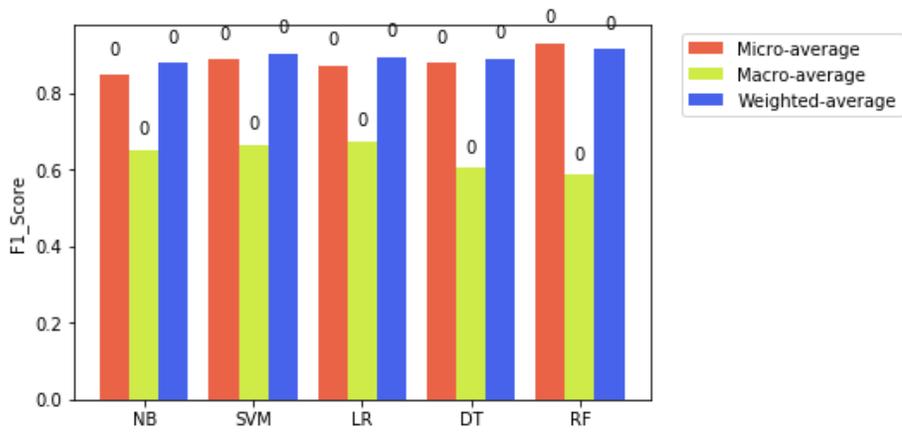

*Figure 6: Barchart of machine learning classifiers' performance based on the Amazon Product Review dataset.*

*Table 11: Results obtained from the machine learning classifiers using the Twitter Sentiment dataset.*

| Twitter Dataset | | | | |
|---|---|---|---|---|
| Classifier | | Precision | Recall | F1 Score |
| Naïve Bayes (NB) | Micro average | 0.6847 | 0.6847 | 0.6847 |
| | Macro average | 0.6843 | 0.6836 | 0.6838 |
| | Weighted average | 0.6845 | 0.6847 | 0.6844 |
| Support Vector Machine | Micro average | 0.6610 | 0.6610 | 0.6610 |
| | Macro average | 0.6607 | 0.6609 | 0.6608 |

|  | Weighted average | 0.6613 | 0.6610 | 0.6611 |
| --- | --- | --- | --- | --- |
| Logistic Regression (LR) | Micro average | 0.6881 | 0.6881 | 0.6881 |
|  | Macro average | 0.6884 | 0.6886 | 0.6880 |
|  | Weighted average | 0.6892 | 0.6881 | 0.6881 |
| Decision Trees (DT) | Micro average | 0.6239 | 0.6239 | 0.6239 |
|  | Macro average | 0.6260 | 0.6254 | 0.6237 |
|  | Weighted average | 0.6270 | 0.6239 | 0.6234 |
| Random Forest (RF) | Micro average | 0.6227 | 0.6227 | 0.6227 |
|  | Macro average | 0.6301 | 0.6261 | 0.6208 |
|  | Weighted average | 0.6316 | 0.6227 | 0.6199 |

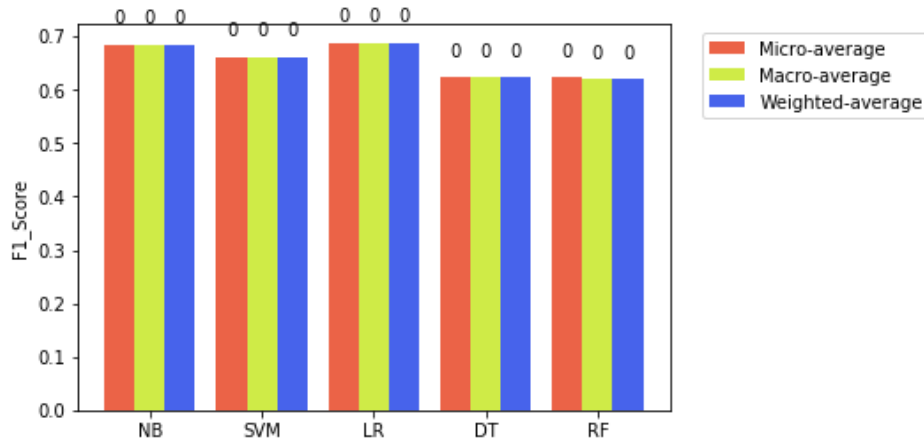

*Figure 7: Barchart of machine learning classifiers' performance based on the Twitter sentiment dataset.*

### 4.2.3 Discussion of Results

The results of these experiments in Tables 9, 10 and 11 show that performance in the classification tasks of a machine-learning algorithm to an extent depends on the number of datasets available. The machine learning algorithms on the IMDB datasets achieve greater prediction accuracy than the other datasets. Also, SVM performs better than the other traditional and non-semantic methods.

The results obtained from the operation of the machine learning algorithms for the Amazon Product Review datasets show a significant difference between the Micro-average and Macro average, this is as a result of the unbalanced data sets. Although a resampling process (upsampling) was carried out the classification algorithms made better predictions with the larger class (positive reviews). The micro-average results of the Random Forest Classifier outperform that of the SVM classifier although the latter's macro-average significantly outperforms the former. With further work, we can perform experiments with balanced datasets to re-evaluate their performance. The bar charts in Figures 5,6 and 7 helps in visualizing the differences between these evaluation types.

The results obtained from the operation of the classifiers on the Twitter Datasets further supports the strong correlation between classification accuracy and the number of data samples. However, the classifier with the best results is the Logistic Regression Classifier and following that is the Naïve Bayes algorithm. This shows promising results as the size of the Twitter datasets are relatively small. This discovery calls for further investigation in developing methods that can harness the strengths of various algorithms to achieve optimal accuracy.

## 4.3 Recurrent Neural Networks – Long- and Short-Term Memory

### 4.3.1 Data Preparation

Recurrent Neural Networks with LSTM layers were implemented in this experiment to demonstrate the use of semantics in sentiment analysis. To implement the textual semantic representations of words, word embeddings were used. Dense vector representations were used to train our semantic models. In this experiment we utilized two types of word embeddings namely:

**Word2Vec**

Word2Vec is a vector representation of words, that either encode the meaning of a word to other words in the same context or the semantics of a context as it relates to the word (Mikolov, Sutskever, Chen, Corrado, & Dean, 2016). In terms of structure, Word2Vec is a shallow, two-layer neural network which is trained to reconstruct the linguistic context of words. The input to this network can be a large corpus of textual data, the output is a vector space typically of several hundred dimensions with each word in the corpus assigned a corresponding vector in the space. Words that share common context are more likely going to be clustered together. Word2Vec is computationally-efficient as a predictive model for learning word embeddings from raw text. There are two variations of Word2Vec namely the Continuous Bag-of-Words (CBOW) and the Skip-Gram model. Figure 8 shows a diagrammatic representation of these models.

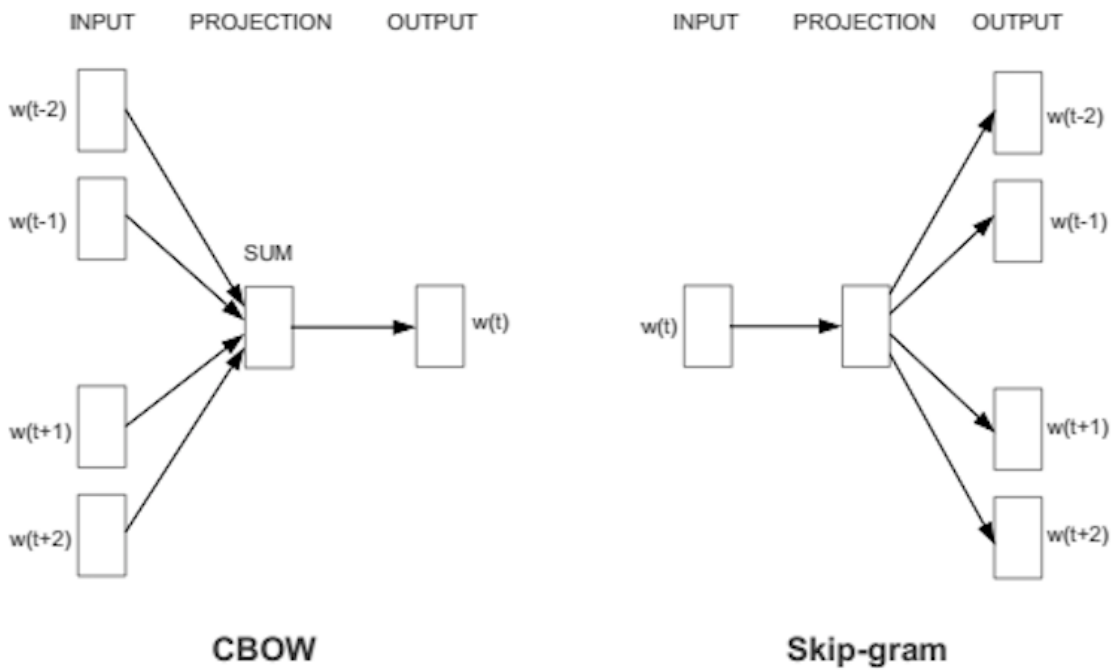

*Figure 8: Diagram of the structure of CBOW and the Skip-gram model*
*Source:* (Mikolov et al., 2016)

**Global Vectors (GloVe)**

This is a model for distributed word representation. It is an unsupervised learning algorithm for obtaining vector representations for words. It uses semantic similarity to map words to a meaningful space between each word. The model is trained based on an aggregated global word-word co-occurrence statistic from a corpus. The resulting representations showcase an interesting linear substructure of the word vector space.

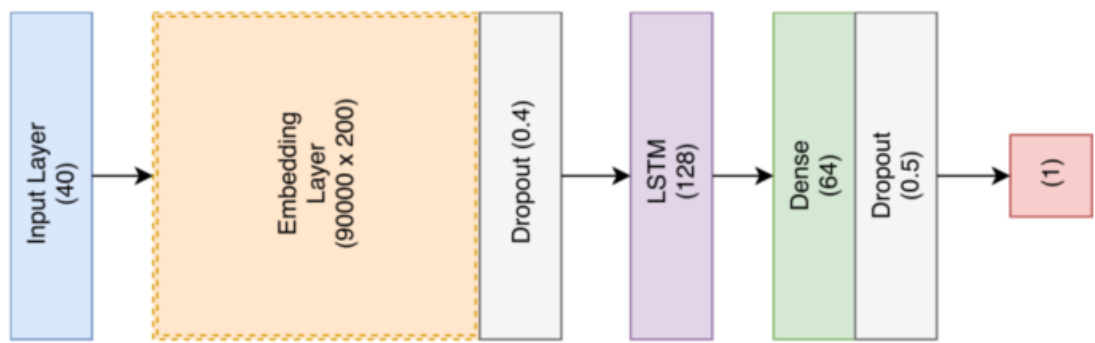

*Figure 9: LSTM Architecture Implemented for the text classification task*

In building the neural network classifier Figure 9 shows the architecture that was implemented with the deep learning framework known as TensorFlow.

*Table 12: Hyper-parameters of the LSTM implemented*

| Hyperparameter | Hyperparameter implemented | Remarks |
| --- | --- | --- |
| Optimizer | Adam Optimizer | Gave us the highest accuracy |

| Loss Function | Binary Cross-Entropy loss | Most suitable for Binary classification tasks |
| --- | --- | --- |
| Epochs | 20, 25, 30 | Varies for the dataset used |
| Batch Size | 50, 100, 150 | Varies for the dataset used |

To improve the performance of the neural network hyper-parameter optimizations were carried out. Table 12 shows the parameters that were used. For the Twitter dataset, we used 20 epochs, for the Amazon product reviews 25 epochs were used, while the IMDB dataset 30 epochs were used. This batch sizes used for the different datasets also follows the aforementioned order.

### 4.3.2 Results

*Table 13: Summary of the Results of the LSTM implemented*

| Dataset | Classifier | Feature Representation | Accuracy |
| --- | --- | --- | --- |
| IMDB Movie Review | LSTM | Word2Vec | 88.64% |
| | | GloVe | 89.12% |
| | LSTM | Word2Vec | 98.18% |

| Amazon Product Consumer Review | | GloVe | 97.44% |
|---|---|---|---|
| Twitter Dataset | LSTM | Word2Vec | 89.82% |
| | | GloVe | 91.20% |

Comparing the best performing algorithms of the non-sematic methods and methods that involve semantics, in most cases, the semantic feature extraction significantly outperforms other non-semantic methods. With further hyper-parameter tuning, the classification processes made on the IMDB dataset may prove this hypothesis to be true.

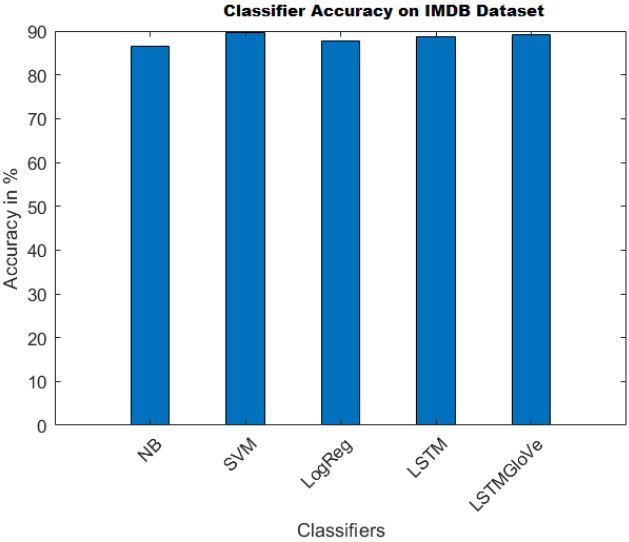

*Figure 10:Sentiment Analysis performance comparison of Semantic and Non-semantic approach based on the IMDB movie review datasets.*

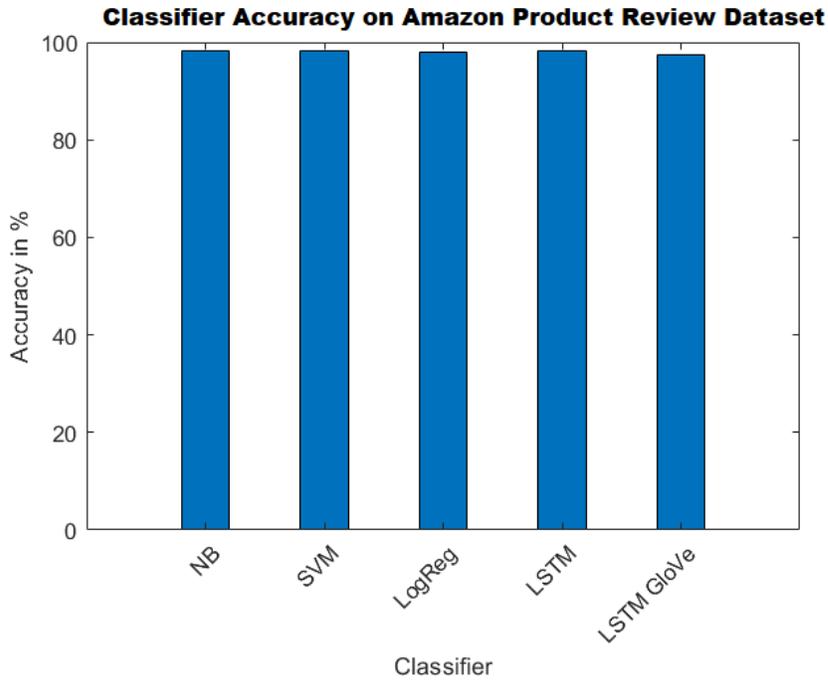

*Figure 11: Sentiment Analysis performance comparison of Semantic and Non-semantic approach based on the Amazon Consumer product review datasets.*

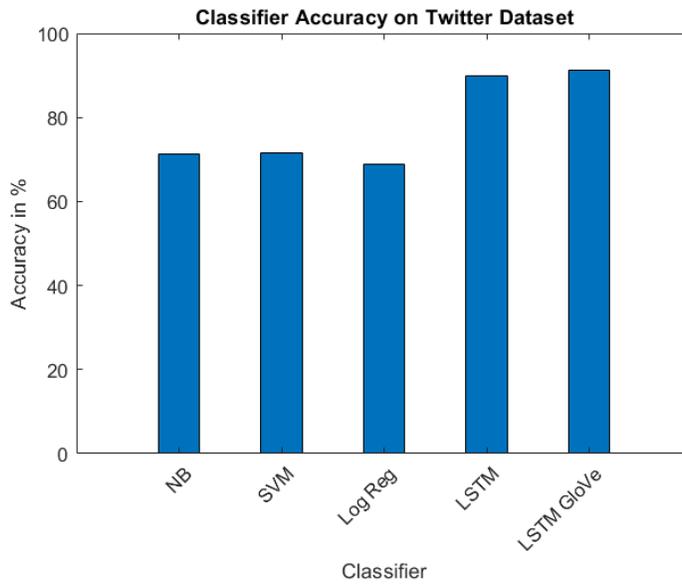

*Figure 12: Sentiment Analysis performance comparison of Semantic and Non-semantic approach based on the Twitter dataset.*

### 4.3.3 Discussion of Results

From the results in Table 12, it can be seen that the LSTM when applied to GloVe, produced the best results. It is worthy of note that one of the reasons why such results were produced is that GloVe tends to encode a better level of semantics when compared to Word2Vec embeddings. In this experiment, an external word embedding was used. The GloVe with 6 billion tokens with a dimension of 100 was utilized to carry out the sentiment analysis task. This embedding was trained using the Wikipedia corpus. With these embeddings, the neural network was trained and used to make classifications based on the semantic encodings with the word embeddings. Further, the bar charts in Figures 10,11 and 12 helps to visualize how well the neural network performs when compared to the other machine learning classifiers.

### 4.4 Further Discussions

To further analyze these findings the best results obtained for each classifier on each dataset are shown in Table 14.

*Table 14: Summary of results obtained from classifiers used in experimentation.*

| Classifier | Dataset | | |
|---|---|---|---|
| | IMDB Movie Reviews | Amazon Product Reviews | Twitter dataset |
| Bayesian Network (BN) | 85.80 | 74.20 | 66.50 |
| Logistic Regression (LR) | 89.69 | 89.42 | 68.81 |
| Support Vector Machine (SVM) | 89.99 | 90.15 | 66.11 |
| Naïve Bayes (NB) | 85.95 | 88.72 | 68.44 |
| Decision Trees (DT) | 71.68 | 88.99 | 62.34 |
| Random Forest (RF) | 74.02 | 91.33 | 61.99 |
| RNN (LSTM) | 89.12 | 98.18 | 91.20 |

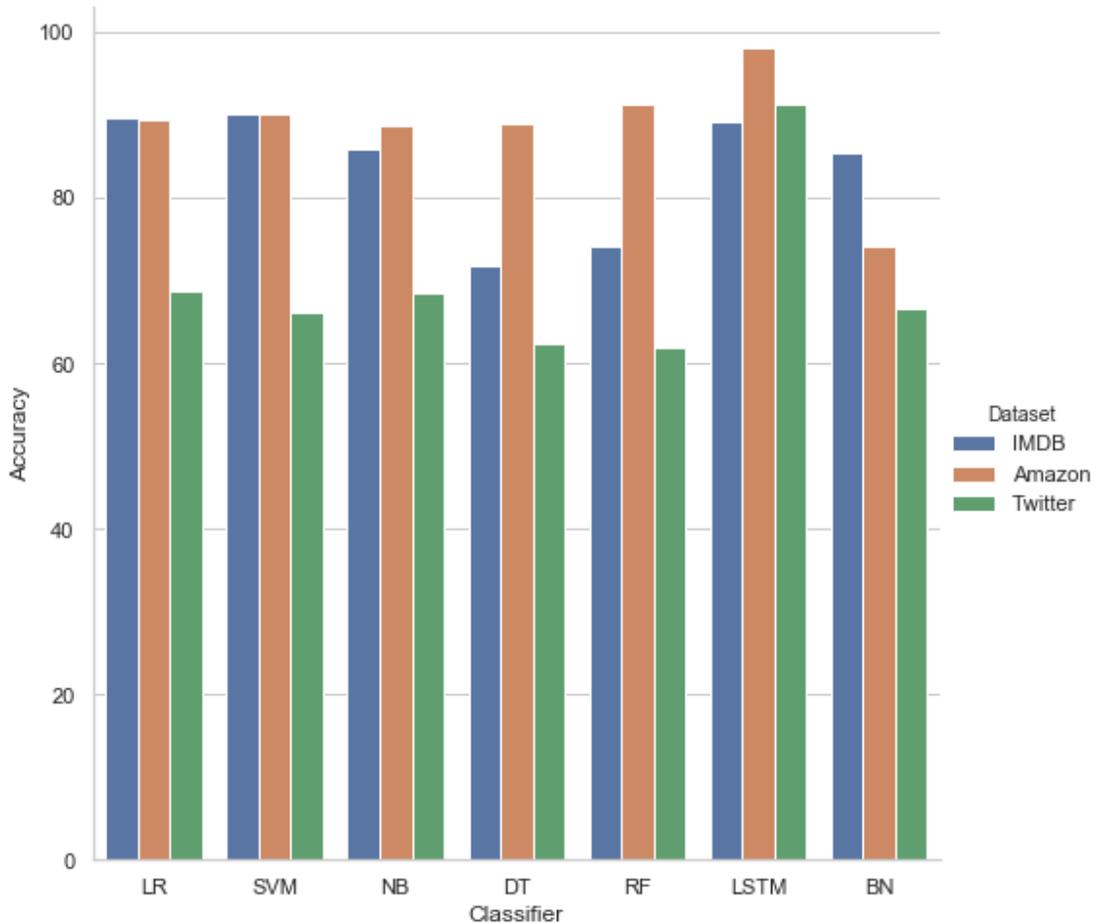

*Figure 13: Bar chart showing the results of the classifiers used in the experiment.*

The results in Table 13 show that Graphical Models like Bayesian Networks also give reasonable results when being used to perform sentiment analysis tasks. The bar chart in Figure 13 helps to visualize these results. Also, it outperforms some traditional machine learning algorithms in a certain task, and it is also robust to overfitting. Although the methods that encode semantics (RNN-LSTM) do not consistently outperform certain non-semantic methods, it is plausible to say that efficient methods of text feature representation possess a great deal of effect in the performance of a classifier. As seen in Table 13, the SVM consistently provides one of the best results because it can produce an efficient separation of classes when features are well represented using vectors. Effective and efficient text representation for Graphical models that can encode semantics can somewhat improve the way they perform text classification tasks. Further study on how the inclusion of semantics will not only be included in the scoring or learning algorithm but also on the text feature representation may improve the performance of graphical models for the sentiment analysis task.

## 4.5 Comparative Analysis of Results

**Bayesian Network:** From the results shown in Table 14 it can be seen that the use of Bayesian Network (BN) produced comparable results to the results obtained by other machine learning classifiers. The results obtained by this model was dependent on the number of variables present in the dataset during classification. These attributes were used by the BN to perform the text classification. Therefore, the number of variables present in the dataset determined the accuracy of the model and to perform this classification, dependencies were created between words and this modeled how the presence of a variable would determine the class of a document. However, an increasing number of attributes will introduce adverse effects on classification accuracy. In other words, obtaining the right attributes to use in a text classification task can be daunting. Therefore, novel methods for attribute selection will need to be developed as this can further improve the performance of the BN model. Obtaining the right number of attributes will also ensure that the model does not overfit. This implies that the ability to obtain an adequate number of attributes will not only improve model performance but will also prevent overfitting as the "most important" attributes are being effectively utilized by the algorithm to perform text classification.

In addition, the time spent on building the Bayesian classifier largely depends on the number of attributes and the learning algorithm and in most cases, the resulting performance produced by the other learning algorithms does not differ significantly; this is evident in Table 6,7 and 8.

**Support Vector Machine (SVM):** The result in Table 14 shows that the SVM performs well for the text classification task. This is true for the following reasons. SVM works well with data represented in High dimensional input space. Most text classification problem can posses' attributes as high as 10,000 attributes. SVM possesses an overfitting protection mechanism so a large number of attributes is not an issue. Secondly, the number of irrelevant features in most text classification problems are few. This implies that well-designed attribute selection techniques the SVM would still perform well with little or no attribute selection techniques. However, one has to be careful in defining a threshold for "relevance" as this might rule out attributes that the algorithm may consider relevant. Thirdly, in using the TF-IDF method of feature representation, a sparse representation of the documents or data is formed. SVM is well suited for problems with sparse representations. Lastly, most text classification problems are linearly separable. So, the SVM is capable of finding these linear separators. Nevertheless, SVMs may not perform well in complex situations were a text or document shows sarcasm. SVMs, as discussed above, are not semantic and are highly dependent on the textual representations. Also, the training time required for SVMs is roughly comparable to Decision Trees and Random Forest algorithms but they are more expensive than the Naïve Bayes and Logistic Regression algorithms.

**Logistic Regression:** In performing text classification problems the LR can suffer from overfitting especially when sparse vector representations like the TF-IDF are used. It does not also perform well with an inadequate number of features. Although regularization techniques can guard it against overfitting. Feature selection is key for its effective application as an excessive number of features can cause overfitting.

**Decision Trees and Random Forest**: Although this model is not the best classifier, this model provided a level of interpretability on how it works. This facilitated attribute selection for the other classifiers. In cases where a single Decision Tree may begin to overfit, the Random Forest can overcome. For the sentiment analysis task, it was discovered that the RF and DT were susceptible to small perturbations and for this removal of stopwords was effective. This classifier is also prone to overfitting, however with the pruning techniques a better performance can be obtained.

**Recurrent Neural Networks (LSTM):** This classifier was able to capture to some extent the semantic and syntactic features of the textual sentiment. This is one of the reasons why it consistently outperforms other classifiers in most cases. However, certain limitations still occur when using this classifier. It performs badly when the datasets available are little. It also suffers from a lack of interpretability as we can not certainly ascertain how the classification is being performed by the algorithm. Unlike the SVM which does not require a lot of hyper-parameter tuning a lot of tuning has to be made to get the best out of the application of this technique.

# CHAPTER FIVE

# CONCLUSION AND RECOMMENDATIONS

## 5.1 Conclusion

In this work, sentiment analysis (a text classification problem), its usefulness and applications in our lives were pointed out. The need for the improvement of this task led to several investigations of how well various machine learning classifiers can be used to carry out this task. As these machine learning classifiers show comparable results, a proper investigation of this work centered on the use of semantics to perform the sentiment analysis task. This work pointed out how Graphical models and neural networks encode semantics in their various methods. From the results obtained, it is safe to say that the better performance of graphical models can be obtained if the use of semantics can be encoded in the text feature representation and its learning algorithms. This method of approach can closely mimic the technique applied in neural networks. For future work, further investigations will need to be carried on the use of ensemble machine learning algorithms for including semantic and non-semantic methods in the ensemble to harness the strength of both algorithms; also further theoretical demonstrations for the establishment of results and exploring the use of graphical models and deep learning networks may need to be worked upon.

## 5.2 Recommendations for Future Work

The experiments performed in this study focused on exploring pieces of evidence related to these questions:

1. Does the inclusion of semantics improve a sentiment analysis or text classification task?

    To further reiterate, probabilistic graphical models can create dependencies between words based on the way they appear in a textual document. This feature representation by graphical models is considered semantic as the company a word keeps can give a level of information to the meaning of the word. However, from the results obtained in the experiments, we observed that in some cases, some of the non-semantic methods outperforms the semantic method of sentiment classification executed by the Bayesian Network. Nevertheless, recurrent neural networks (long-and-short-term-memory neural networks) come to the rescue as not only is the notion of semantics encoded in the text representation (Word2Vec and GloVe) but also the network to an extent can extract syntactic and semantic features. With this discovery, it is important to suggest that further work on the encoding of semantics on the textual representation of documents well suited for probabilistic graphical models(PGMs) and the development of novel graph learning algorithms for PGMs will need proper investigation as

       this can bring us closer to the technique being employed by the neural network.

2. Does the creation of word dependencies between words in a sentence created by Probabilistic Graphical models help tackle the issue of semantic ambiguity inherited in the documents?

       The creation of word dependencies between words that are related in meaning is obtained by how they are used in textual documents. With appropriate model training methods, PGMs can create word dependencies between words that show semantic ambiguity and can attempt to make predictions based on the dependencies created. However, these novel algorithms that can train models to learn semantic ambiguity will need further investigation.

To further establish our hypothesis explanations based on theoretical experiments can further explain how the intentional inclusion of semantics can improve text classification accuracy. With this, further research can be done to carry a more fine-grained level of classification with more than two classes. For instance, we should be able to measure how certain words in a document inform the classification decision made by the algorithm for more classes like positive, neutral and negative. If we can measure the degree of dependence of a group of words in a sentence and how it improves the probability of algorithm to make the right classification, this would a long way to tackle more challenging problems in sentiment analysis.

Secondly, we can further explore ensemble methods, in this project, all the classifiers used in this project were standalone (except for the Random Forest classifier), further experiments can be done to integrate more than one classifier. Ensemble methods have previously shown promising results; further implementation of this method can be a way to harness the strengths and mitigate the weaknesses inherent in these classifiers.

Further variations of PGMs can be used, the combination of graphical models and neural networks is one area that may prove to show promising results in this task. A theoretical explanation and empirical evaluation of this method may be promising and should be considered further investigations.

*Subseries Lecture Notes in Artificial Intelligence and Lecture Notes in Bioinformatics)*, *4724 LNAI*, 501–511. https://doi.org/10.1007/978-3-540-75256-1_45

Demers, E., & Vega, C. (2008). Soft Information in Earnings Announcements : News or Noise? *International Finance Discussion Paper*, *2008*(951), 1–56. https://doi.org/10.17016/ifdp.2008.951

Driver, E., & Morrell, D. (1995). Implementation of Continuous Bayesian Networks Using Sums of Weighted Gaussians. *Proceedings of the Eleventh Conference on Uncertainty in Artificial Intelligence*, 134–140.

Fan, R. E., Chang, K. W., Hsieh, C. J., Wang, X. R., & Lin, C. J. (2008). LIBLINEAR: A library for large linear classification. *Journal of Machine Learning Research*, *9*(2008), 1871–1874. https://doi.org/10.1145/1390681.1442794

Friedman, N., & Goldszmidt, M. (1996). Discretizing Continuous Attributes While Learning Bayesian Networks. *Icml*, 157–165. https://doi.org/10.1001/archinte.159.12.1359

Genkin, A., Lewis, D. D., & Madigan, D. (2007). Large-scale bayesian logistic regression for text categorization. *Technometrics*, *49*(3), 291–304. https://doi.org/10.1198/004017007000000245

Giovanelli, C., Liu, X., Sierla, S., Vyatkin, V., & Ichise, R. (2017). Towards an Aggregator that Exploits Big Data to Bid on Frequency Containment Reserve Market. *Proceedings IECON 2017 - 43rd Annual Conference of the IEEE Industrial*
78

Ravi, K., & Ravi, V. (2015). A survey on opinion mining and sentiment analysis: Tasks, approaches and applications. *Knowledge-Based Systems*, *89*, 14–46. https://doi.org/https://doi.org/10.1016/j.knosys.2015.06.015

Ren, F., & Kang, X. (2013). Employing hierarchical Bayesian networks in simple and complex emotion topic analysis. *Computer Speech and Language*, *27*, 943–968. https://doi.org/10.1016/j.csl.2012.07.012

Ren, F., & Quan, C. (2012). Linguistic-based emotion analysis and recognition for measuring consumer satisfaction: an application of affective computing. *Information Technology and Management*, *13*(4), 321–332. https://doi.org/10.1007/s10799-012-0138-5

Rustamov, S., Mustafayev, E., & Clements, M. A. (2013). *Sentiment Analysis using Neuro-Fuzzy and Hidden Markov Models of Text*. (March). https://doi.org/10.1109/SECON.2013.6567382

Saif, H., He, Y., & Alani, H. (2012). *Semantic Sentiment Analysis of Twitter*. https://doi.org/10.1007/978-3-642-35176-1_32

Sehgal, D., & Agarwal, A. K. (2018). *Real-time Sentiment Analysis of Big Data Applications Using Twitter Data with Hadoop Framework BT - Soft Computing: Theories and Applications* (M. Pant, K. Ray, T. K. Sharma, S. Rawat, & A. Bandyopadhyay, Eds.). Singapore: Springer Singapore.

Snell, D. R. C. E. J. (2018). *Analysis of Binary Data, Second Edition* (Second edi).

3-319-18123-3

Yin, S., Han, J., Huang, Y., & Kumar, K. (2014). Dependency-Topic-Affects-Sentiment-LDA Model for Sentiment Analysis. *Proceedings - International Conference on Tools with Artificial Intelligence, ICTAI*, *2014-Decem*, 413–418. https://doi.org/10.1109/ICTAI.2014.69

Zhang, K., Downey, D., Chen, Z., Cheng, Y., Xie, Y., Agrawal, A., … Choudhary, A. (2013). A probabilistic graphical model for brand reputation assessment in social networks. *Proceedings of the 2013 IEEE/ACM International Conference on Advances in Social Networks Analysis and Mining, ASONAM 2013*, 223–230. https://doi.org/10.1145/2492517.2492556

Zhang, K., Xie, Y., Yang, Y., Sun, A., Liu, H., & Choudhary, A. (2014). Incorporating conditional random fields and active learning to improve sentiment identification. *Neural Networks*, *58*, 60–67. https://doi.org/10.1016/j.neunet.2014.04.005

Zhao, X., & Ohsawa, Y. (2018). *Sentiment Analysis on the Online Reviews Based on Hidden Markov Model*. https://doi.org/10.12720/jait.9.2.33-38